\documentclass[letterpaper, 10 pt, conference]{ieeeconf}  

\IEEEoverridecommandlockouts                              

\usepackage{amsmath}   
\usepackage{amssymb}   
\usepackage{booktabs}  
\usepackage{multirow}  
\usepackage{makecell}  
\usepackage{threeparttable}  
\usepackage[font=footnotesize,labelsep=colon,justification=justified]{caption}  
\usepackage{algorithm}   
\usepackage[noend]{algorithmic}
\usepackage{todonotes}
\newcommand{\sysname}{GenGenIK}
\renewcommand{\paragraph}[1]{%
  \par\smallskip\noindent\textit{#1}\par\nobreak\noindent\ignorespaces
}
\newcommand{\SE}[1]{\mathrm{SE}(#1)}
\newcommand{\FK}{\mathrm{FK}}

\title{\LARGE \bf
Generalized Inverse Kinematics via Conditional Flow
}

\author{Kaitlyn Zhang, Tom\'as Lozano-P\'erez, Leslie Pack Kaelbling\\
{\normalsize Massachusetts Institute of Technology}\\
{\tt\small kyz@mit.edu \quad \{tlp, lpk\}@csail.mit.edu}
}

\usepackage{stfloats}
\usepackage{graphicx}
\usepackage{hyperref}
\hypersetup{hidelinks}
\usepackage{xcolor}

\DeclareMathOperator{\atantwo}{atan2}

\begin{document}

\maketitle

\begin{abstract}
Inverse kinematics (IK) is a foundational primitive in robotics, mapping the task-space objectives expressed by higher-level planners to robot configurations that realize them. Traditional IK methods generally treat each query as an isolated pose-to-configuration problem, but planners often need a distribution over valid realizations that accounts for future reachability and preferences such as collision avoidance and stability. We introduce \sysname, a generative modeling approach that supports mobile bases, multiple end-effectors and targets, scene awareness, and preferences over robot behavior within a common interface. Experiments show that \sysname{} outperforms existing methods in both solve rate and runtime across tasks such as bimanual lift and flip-up, while allowing flexible control over the sampled distribution and remaining effective under partial scene observations. 
Website: \url{https://icra-2027-gengenik.github.io/}.
\end{abstract}

\section{INTRODUCTION}
Many robot operations are most conveniently described in Cartesian end-effector
coordinates: for example, opening a door, grasping or pushing an object on a
table, pouring liquid, etc.  However, safety verification, motion
planning, and many control processes require a full configuration space (C-space)
realization of the robot. The problem of generating valid joint configurations that satisfy constraints on the end-effector is known as the {\em inverse kinematics} (IK) problem. For redundant robots, in which the dimension of the C-space exceeds the dimension of the end-effector space, the IK problem is computationally difficult, and most often addressed by non-linear optimization or other numerical methods.

When planning for long-horizon manipulation tasks, it is generally necessary to
consider many alternative IK solutions to individual steps, for example,
considering different base positions for opening a door or reaching into a
cupboard~\cite{reachinv}.  Numerical methods for IK generate single solutions; they can
typically be guided into generating multiple solutions but that process can be slow and may not reliably cover the feasible solution space.

For highly redundant robots, such as humanoids, it is not
sufficient to generate samples of configurations consistent with an
end-effector constraint.  We will typically have a large number of additional
constraints and preferences on the resulting configuration: the robot should be
stable, the hand should be in the field of view of the camera, multiple
end-effector positions should be reachable from the same base position, the robot should
not be in self-collision or in world-collision, etc.  Generating large
numbers of solutions and rejecting them on the basis of the constraints is not
practical.  Numerical IK solvers can accept a rich set
of constraints and preferences, but as indicated earlier, are slow when multiple
solutions are required.

In this paper we address the {\em generalized IK} problem.  We assume the robot
has a base and a set of end-effectors, such as hands or posable cameras.  
We specify a sequence of targets for the end-effectors, scene collision geometry,
and a set of preferences, such as high manipulability.  The desired result
is a set of whole-body configurations for each of the targets that achieve the
specified end-effector pose and which approximately optimize the preferences.

We model generalized IK as conditional generation over whole-body configurations
or trajectories. We learn a Transformer model~\cite{transformer}, \sysname, based on a masked representation
that lets a single learned model handle different numbers of targets and end-effectors, mobile bases, and partial configuration or trajectory completion
without changing the architecture. Our contributions are:
\begin{enumerate}
        \item \textbf{Generalized IK as conditional C-space completion:} a
          unified formulation encompassing IK over multiple targets,
          multiple end-effectors, mobile bases, warm starts, and partial configuration or
          trajectory infill. 
        \item \textbf{Scene-conditioned generation:} a generative solver that
          directly conditions the distribution on scene geometry, concentrating
          samples on collision-free regions of solution space rather than
          relying on scene-agnostic proposals followed by online search or
          optimization. 
        \item \textbf{Realization under constraints and preferences:} a method for incorporating constraints and preferences via {\em auxiliary training losses}, such as keeping the end-effector in the robot's view; {\em guided sampling} for differentiable constraints, such as manipulability and static stability; and {\em preference conditioning}, which biases the sampled distribution according to non-differentiable requests such as torso use and arm posture.
\end{enumerate}

\begin{figure}[t]
  \centering
  \includegraphics[scale=0.16]{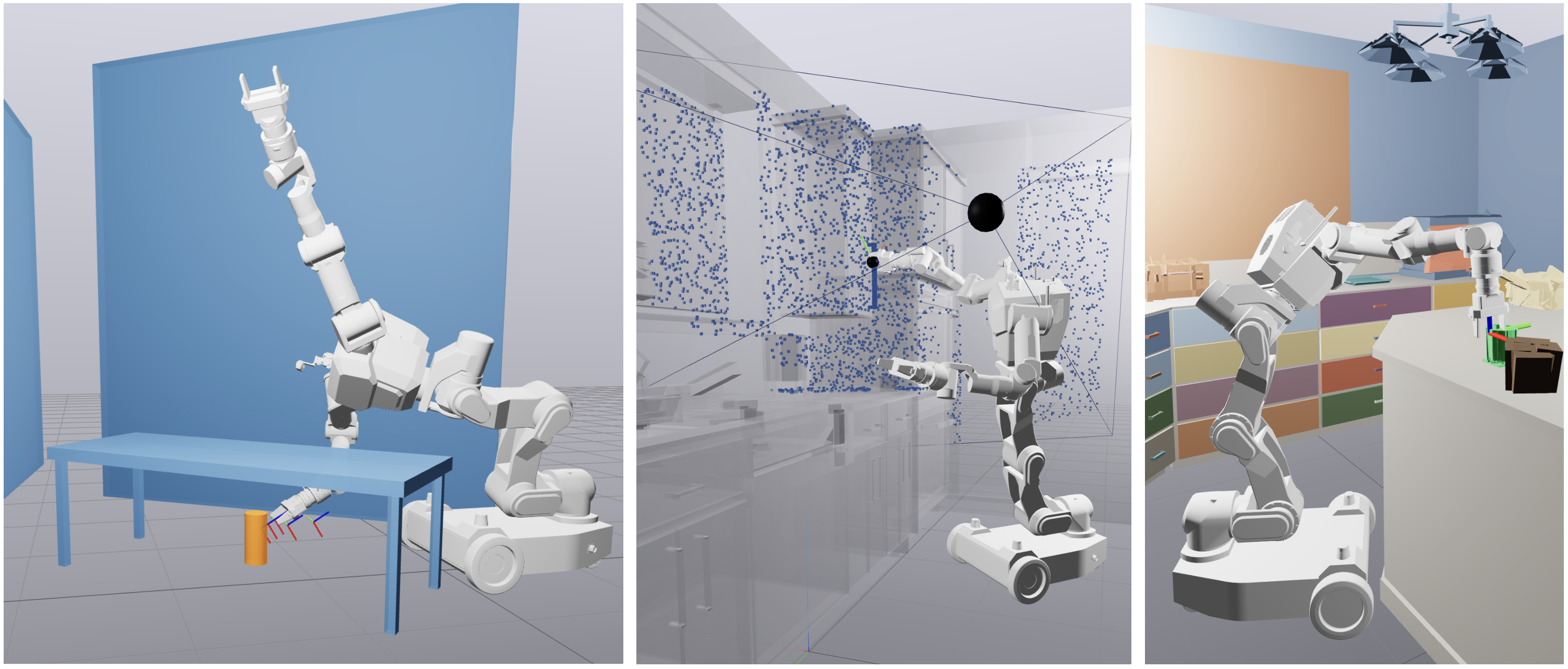} 
  \caption{Example use cases of \sysname{} (under-table pickup, partially observable scenes, cluttered kitchens).}
  \label{fig:scenes}
\end{figure}


\section{RELATED WORK}

\paragraph{Classical inverse kinematics}
Analytic IK solvers such as IKFast~\cite{ikfast} can efficiently enumerate
closed-form solutions when the kinematic structure admits them, but redundancy beyond the task dimension must be handled by sampling free parameters rather than directly representing the full constrained solution distribution.  Numerical methods such as those
in Drake~\cite{drake} and cuRobo~\cite{curobo} support redundant
robots and richer constraints, including self- and world-collision constraints,
multiple task objectives, and whole-body posture costs, but generating multiple
distinct solutions generally requires repeated solves from different
initializations.

\paragraph{Generative inverse kinematics}
Learned IK methods amortize this repeated search by modeling the solution set
directly. IKFlow~\cite{ikflow} uses a conditional normalizing flow to rapidly
generate diverse solutions for a desired end-effector pose, while
IKDiffuser~\cite{ikdiffuser} extends generative IK to more general kinematic
trees with multiple end-effectors and supports differentiable preferences. These methods demonstrate that learned
samplers can efficiently represent multimodal IK solution sets, but primarily
consider single IK targets in free space.  

\paragraph{Scene-aware and pathwise IK}
Other work extends learned or sampled IK to environmental and pathwise
constraints. ViIK~\cite{viik} conditions IK generation on visual scene
observations, while CppFlow~\cite{cppflow}, IKLink~\cite{iklink}, and
STAMPEDE~\cite{stampede} address pathwise IK by selecting or optimizing
configurations across task-space targets. 

\sysname{} efficiently generates diverse solutions to the union of these problems, handling environment
and path constraints while also supporting mobile bases, multiple end-effectors, and
behavioral preferences.


\section{PROBLEM SETTING}

Let $R$ denote a robot embodiment with $E$ end-effectors and whole-body configuration space $\mathcal{Q}_R$, and
let $Q = (q_1, \ldots, q_L) \in \mathcal{Q}^L_R$ denote a whole-body
configuration trajectory of $L \geq 1$ targets. 
End-effector poses are given by the known forward kinematics of the robot $\mathrm{FK}_R(q)$.
The generalized IK task is, given a sequence of vectors of end-effector targets, $H \in \SE{3}^{L\times E}$, to produce a set of {\em realizations}, each of which is a trajectory of full-body configurations $Q \in \mathcal{Q}_R^L$.

We represent a task specification as $T = (H, M)$, where masks $M = (M_T, M_Q)$
are Boolean arrays that specify which end-effector targets must be satisfied and which realization variables must be predicted. This representation allows the same model to
handle different numbers of targets and end-effectors and perform partial configuration or trajectory completion. Let $S$ denote a geometric observation of the scene, represented in our experiments as a robot-centric point cloud or a set of geometric primitives, and let $F(S,q)$ be true iff $q$ is collision-free with respect to $S$.

For a given robot $R$, task $T$, and scene $S$, a realization $Q$ is feasible if every
active target in $T$ is reached, all joint limits are respected, and the robot is free of self- and world-collision: that is, for all $\ell, e$ such that $M_T[ \ell, e] = 1$, $\FK_R(q_\ell)_e \approx H_{\ell, e}$ and for all $q \in Q$, $q \in \mathcal{Q}_R$ and  $F(S, q)$.

For paths, task-space constraints are imposed only at the specified targets. Between targets, we densify the configuration-space interpolation and check physical feasibility, including self- and world-collision and stability. Denote the set of feasible realizations by  
\[\mathcal{F}(R, T, S) = \{Q \in \mathcal{Q}_R^L: Q \text{ satisfies } T \wedge \text{feasible in }S\}.\]

This formulation recovers several familiar IK problems through masking. With
$L=1$, it reduces to classic single target (point) IK; increasing $L$ gives multi-target (path) IK. Activating multiple target entries at a point gives multi-end-effector IK, while
changing $M_Q$ determines which base or joint variables are fixed, generated, or
partially specified.

Generalized IK defines a set of valid realizations rather than a unique
solution. For redundant robots, $\mathcal{F}(R, T, S)$ may contain continuous
families of solutions and multiple disconnected modes. We therefore model
generalized IK as conditional sampling, \[Q \sim p_\theta (Q \mid R, T, S, D),\]
where $\mathcal{F}(R, T, S)$ defines the feasible support and $D$ optionally
specifies preferences over otherwise-valid solutions, such as manipulability,
stability, or preferred whole-body posture. Our goal is to
efficiently sample diverse useful solutions from the feasible set rather than
find only one $Q^* \in \mathcal{F}(R, T, S)$.  

We evaluate all methods through the common
interface
\[\texttt{sample}((R, T, S,D), K) \rightarrow \{Q^{(1)}, \ldots,
  Q^{(K)}\}\] where $K$ is the requested
sample budget. For each candidate set, we examine \textbf{validity} (task-space
error, limits, and collision), \textbf{sample diversity}, \textbf{request satisfaction} for optional realization
objectives, and \textbf{sampling latency}. 
Samples from any generative model may typically be further improved by executing a {\em refiner} that performs additional gradient-based optimization. We measure both the pre- and post-refinement quality of the samples produced.


\section{METHOD}

We realize $p_\theta(Q \mid R, T, S, D)$ as a masked conditional flow-matching
model~\cite{flowmatching} over a feature encoding of $Q$. The model is trained on procedurally
generated feasible solutions and sampled via few-step inference, optionally
followed by redraw or numerical refinement. 

\subsection{Data generation}
\label{sec:method-data}

Each training example is stored as a problem-solution tuple $(Q, R, T, S, D)$. 
For scene-free point IK, we sample $q$
uniformly and compute targets by FK. 
More generally, we sample a scene $S$ and a task $T$ and use cuRobo~\cite{curobo} to generate candidate solutions. All retained examples are independently certified with Drake~\cite{drake} mesh geometry for the applicable self- and scene-collision constraints.
The solver additionally includes costs for target visibility or robot stability when these solution attributes are desired. 
For paths, we solve IK over all base and joint variables at the first target and then fix the base while solving subsequent targets. This trains the model to learn base placements from which
the entire task can be completed without repositioning. We generate data across diverse manipulation tasks and scene distributions, including procedurally generated environments and more realistic simulated household scenes. 

We further vary how each generated solution is presented during training through 
a mask policy that determines which end-effector values are constrained and which $q$ coordinates are 
fixed versus generated. The policy matches conditioning patterns expected at inference,
so the model learns distributions over free $q$ coordinates given fixed ones and over unconstrained end-effector targets given constrained ones. 
To train the model to condition on optional solution preferences ($D$), we use hindsight-style relabeling~\cite{hindsight_relabeling}, pairing each stored solution with sampled requests that it already satisfies.

\paragraph{Featurization}
\label{sec:featurization}
Robot configurations and task targets are mapped to a common representation
before being passed to the model. A solution $Q$ with $L$ targets is represented as $Y \in \mathbb{R}^{d_b} \times \mathbb{R}^{L \times d_q}$,
with one base block of dimension $d_b$ and one joint-state block of dimension $d_q$ per target. We normalize interval-bounded joints as
\[\tilde{q}_d = \frac{q_d - \bar{q}_d}{\rho_d} \in [-1, 1],\]
where $\bar{q}_d$ and $\rho_d$ are the midpoint and half-width of the joint
range. For these bounded coordinates, we pass model outputs through an
interior-identity soft bounding function
\[\mathcal{B}_d(x) = \begin{cases} x, & |x| \leq a_d, \\ \text{sign}(x)\!\left[
a_d + m_d \tanh\!\left(\frac{|x|-a_d}{m_d}\right) \right], & |x| > a_d, \end{cases} \]
where $m_d = \mu / \rho_d$ and $a_d = 1-m_d$ for joint $d$, and
$0 < \mu < \rho_d$ is a specified safety margin. This function leaves values
unchanged away from the limits and smoothly saturates toward $(-1,1)$ near them.
We represent continuous revolute joints and mobile-base heading with $(\cos q, \sin q)$ pairs and use direct coordinates for planar base position. We use the function \textit{decode} to map vectors back to configurations.

We represent end-effector targets using translation and the first two columns of
the rotation matrix, yielding nine features per target~\cite{rot6d}. 

We express each problem in a canonical task frame defined by the first
observed end-effector target. We remove its planar translation and yaw from the
robot base, the remaining task targets, and the scene point cloud, making the
representation invariant to global planar translation and rotation. Remaining
continuous features are standardized using training-set statistics, while
naturally bounded $[-1, 1]$ coordinates retain their original scale.  

\subsection{Conditional flow matching}
\label{sec:method-training}
There are many approaches to conditional generative modeling, including CVAEs~\cite{cvae} and diffusion models~\cite{ddpm}. We use {\em conditional flow matching} because it can model the multimodal solution distributions induced by IK while generalizing diffusion-style probability paths to more direct transport paths that support efficient few-step sampling~\cite{flowmatching}.
We generate robot realizations $Q$ conditioned on the problem specification, where $c$ denotes the encoded conditioning information derived from $(T, S, D)$ and the free-coordinate mask $M_Q$ specifies which coordinates of $Q$ are generated. 

We train a network to transport a Gaussian source distribution to the
conditional distribution of feasible
solutions~\cite{flowmatching,rectified_flow}; Algorithm~\ref{alg:training} summarizes the training procedure. For a dataset solution $Q$
with feature encoding $Y$, we sample $Z \sim \mathcal{N}(0, \sigma^2 I), \,\, s
\sim \mathcal{U}(0, 1 - \epsilon),$ and construct the masked straight-line
interpolant
\begin{equation*} 
X_s = M_Q \odot [(1-s) Z + sY] + (1-M_Q) \odot Y
\end{equation*}
where $M_Q=1$ on generated coordinates and $M_Q = 0$ on fixed
coordinates. We sample $s$ only while a minimum remaining flow time $\epsilon$
remains, preventing the denominator in Eq. \ref{eq:velocity} from becoming
arbitrarily small.  

Along this path, the target velocity is \begin{equation*}
  u^\star = M_Q \odot (Y - Z) = \frac{Y - X_s}{1 - s}.
\end{equation*}

\begin{algorithm}[t]
\caption{Conditional flow-matching training}
\label{alg:training}
\begin{algorithmic}[1]
\REQUIRE training pair $(Q, c)$, free-coordinate mask $M_Q$
\STATE Featurize $Q$ as $Y$
\STATE Sample $Z \sim \mathcal{N}(0, \sigma^2 I)$ and $s \sim \mathcal{U}(0, 1 - \epsilon)$  
\STATE $X_s \leftarrow M_Q \odot [(1-s) Z + sY] + (1-M_Q) \odot Y$
\STATE $\hat Y \leftarrow \mathcal{B}\left(X_s + g_\theta(X_s,s,c)\right)$
\STATE $v_\theta \leftarrow (\hat Y-X_s)/(1-s)$
\STATE $\mathcal{L}_{\mathrm{flow}}
    \leftarrow \operatorname{mean}_{M_Q=1}
    \|v_\theta-(Y-Z)\|_2^2$
\STATE Compute $\mathcal{L} = \mathcal{L}_{\mathrm{flow}}+\mathcal{L}_{\mathrm{aux}}$

\IF{few-step fine-tuning}
    \STATE Roll out the $k$-step sampler with gradients
    \STATE Add task-space loss on the decoded solution to $\mathcal{L}$
\ENDIF
\STATE Update $\theta$ via AdamW to minimize $\mathcal{L}$ using $\partial \mathcal{L} / \partial \theta$
\end{algorithmic}
\end{algorithm}

\subsubsection{Bounded endpoint parameterization}

Rather than predicting $u^\star$ directly, the network $g_\theta(\cdot)$ predicts a residual
toward the true solution,
\begin{equation*}
\hat{Y}_\theta (X_s, s, c) = \mathcal{B} (X_s + g_\theta(X_s, s, c)),
\label{eq:endpoint}
\end{equation*}
and uses the joint-limit-safe endpoint to compute the corresponding velocity analytically
\begin{equation}
  v_\theta(X_s, s, c) = \frac{\hat{Y}_\theta (X_s, s, c) - X_s}{1- s}.
\label{eq:velocity}
\end{equation}

The core objective is masked velocity regression:
\begin{equation*}
  \mathcal{L}_\text{flow} = \text{mean}_{M_Q = 1} \|v_\theta(X_s, s, c) - (Y-Z)\|^2_2.\label{eq:flow_loss}
\end{equation*}
Under the endpoint parameterization,
\begin{equation*}
  v_\theta - (Y - Z) = \frac{\hat{Y}_\theta - Y}{1 - s},
\end{equation*}
so the velocity MSE is equivalently endpoint error weighted by $1 / (1 -
s)^2$. Pre-specified coordinates have zero target velocity and are excluded by $M_Q$.

\subsubsection{Auxiliary losses}

Flow matching provides the primary supervision, while optional auxiliary terms
optimize additional objectives. 

\textit{Visibility.} To encourage camera configurations that keep a
designated task point within the field of view of the robot, we penalize
violations of a viewing cone. Let $\hat{v}_\text{cam}(q)$ denote the
camera viewing axis in world coordinates and $\hat{d}(q)$ the unit direction
from the camera origin to the target point. We use the squared hinge
loss \[\mathcal{L}_\text{view} = \mathrm{avg}_q \left[\max \left(0,
      \cos(\theta_\text{fov} - \alpha) - \hat{v}_\text{cam}(q)^\top
      \hat{d}(q)\right)^2\right].\] Here, $\theta_\text{fov}$ is the
field-of-view half-angle and $\alpha$ is a training-only margin; we average over configurations $q \in Q$. The loss is
zero once the target lies within the tightened cone, preserving a range of valid
camera orientations rather than collapsing predictions toward the cone
center.

\textit{$S^1$ consistency.} Continuous revolute joints and base heading are
encoded as $(\cos q, \sin q)$ pairs. Because $\atantwo$ decoding is invariant to
feature magnitude, off-circle predictions introduce an irrelevant degree of
freedom in the generative state space. We therefore regularize circular features
$(u_p, v_p)$ to unit norm with \begin{equation*} 
  \mathcal{L}_{S^1}= \mathbb{E} \left[\left(u_p^2 + v_p^2 - 1\right)^2\right].
\end{equation*}

\subsection{Few-step inference}
\label{sec:method-inference}

We generate samples from the trained model using Algorithm~\ref{alg:sampling}.  For fast
inference, we integrate the learned field with a small number of Euler steps on
a nonuniform time schedule, 
\begin{equation*}
    s_i = 1 - \left(1 - i/k\right)^{\gamma}, \qquad i = 0, \ldots, k
\end{equation*}
where $\gamma > 1$ allocates more steps near $s = 1$, when the predicted
endpoint must resolve fine task-space error. We initialize free coordinates
from $Z$ and keep known coordinates fixed by zeroing
their update at every step. The Euler update is  
\begin{equation*}
X_{i+1} = X_i + (s_{i+1} - s_i)\, M_Q \odot \frac{\hat Y_i - X_i}{1 - s_i},
\label{eq:euler}
\end{equation*}
with $\hat Y_i = \hat Y_\theta(X_i, s_i, c)$. Since $s_k = 1$, $s_k - s_{k-1} = 1 - s_{k-1}$, and therefore \begin{equation*}
  X_k = X_{k-1} + (1 - s_{k-1}) \frac{\hat{Y}_{k-1} - X_{k-1}}{1 - s_{k-1}} = \hat{Y}_{k-1},
\end{equation*}
provided $1 - s_{k-1} \geq \epsilon$: the returned sample is the model's final bounded endpoint prediction.
We select inference hyperparameters $k$ and $\gamma$ on held-out validation problems.

\begin{algorithm}[t]
\caption{Few-step IK sampling}
\label{alg:sampling}
\begin{algorithmic}[1]
\REQUIRE query $c$, free-coordinate mask $M_Q$, steps $k$, time exponent $\gamma$, optional redraw threshold $\tau$ 
\STATE Sample $X_0$ from the Gaussian source on free coordinates; insert known values
\STATE $s_i \leftarrow 1-(1-i/k)^\gamma$, \quad $i=0,\ldots,k$
\FOR{$i=0,\ldots,k-1$}
    \STATE $\hat Y_i \leftarrow
    \mathcal{B}\!\left(X_i+g_\theta(X_i,s_i,c)\right)$
    \STATE $v_i \leftarrow M_Q\odot(\hat Y_i-X_i)/(1-s_i)$
    \STATE $X_{i+1} \leftarrow X_i+(s_{i+1}-s_i)v_i$
\ENDFOR
\STATE Decode $X_k$ to obtain $Q$
\IF{redraw is enabled and $e_{\mathrm{FK}}(Q,c)>\tau$}
    \STATE Draw one additional independent sample $Q'$
    \STATE $Q \leftarrow \arg\min_{\tilde Q\in\{Q,Q'\}} e_{\mathrm{FK}}(\tilde Q,c)$
    \ENDIF
\STATE $Q \leftarrow \textsc{Refine} (Q,c)$
\RETURN $Q$
\end{algorithmic}
\end{algorithm}

\subsection{Sampler-aware fine-tuning}
Flow-matching supervision acts on interpolants of data solutions rather than states produced by the sampler itself. We therefore optionally fine-tune on sampler-generated states, in the spirit of reward fine-tuning through diffusion rollouts~\cite{draft}. 

\textit{Task-space rollout.} For few-step deployment, the coarse discretization departs from the continuous objective, so we differentiate through the full $k$-step schedule and penalize the decoded output's task-space error. 

\textit{Scene collision.} For scene-conditioned models, we add collision penalties that apply only to the model's predicted output. We use differentiable FK to place robot collision spheres at the predicted world coordinates and penalize self-collision and the worst penetration into the scene, using a signed distance field~\cite{mpinets}. We pair these penalties with a task-space loss to prevent the model from trading task-space accuracy for collision avoidance.

\subsection{Reflow distillation} Few-step inference with rollout fine-tuning works well at small $k$, but reducing inference to a single network evaluation requires further compression of the learned flow. We therefore use reflow~\cite{rectified_flow, instaflow} to distill a converged multi-step teacher into a one-step student. For each training query $c$, the teacher maps a latent $Z$ to 
\begin{equation*}
    Y^{\text{teach}} = \text{sample}_{k_\text{full}} \left(Z; c, \theta_\text{teach}\right),
\end{equation*} and the student is trained on the resulting $(Z, Y^{\text{teach}})$ pairs, trading task-space accuracy for lower inference cost. 

\subsection{Guided sampling}
\label{sec:method-guidance}
We can shift the learned distribution at test time in two ways. For a warm start $Q^\text{prior}$, we start partway along the flow from a blend of $Z$ and the encoded prior, then integrate the remaining flow to the output. For differentiable objectives, we guide sampling using a task energy.

Following gradient-based guidance for generative inverse problems~\cite{dps} and IK~\cite{ikdiffuser}, we evaluate the energy on the predicted output and differentiate through decoding and robot kinematics. At generation step $i$, let $\hat{Y}_i$ denote the model's predicted raw
output. Given an energy $E(Q)$, we can let \begin{equation*} 
  \hat{Y}_i' = \hat{Y}_i - \eta \nabla_{\hat{Y}_i} E \left(\text{decode}(\hat{Y}_i)\right),
\end{equation*}
and replace the model velocity with \begin{equation*}
  v_i' = \frac{\hat{Y}_i' - X_i}{1 - s_i}
\end{equation*}
in the usual Euler update.

\subsubsection{Guidance objectives} Guidance energies may be combined with a weighted sum $E(Q, s) = \sum_j w_j \omega_j (s) E_j(Q)$, 
where $w_j$ controls the objective strength and $\omega_j(s)$ activates the
objective over select portions of the generation trajectory. 
We consider the following objectives, and others can be defined analogously.

\textit{Manipulability:} To discourage singular configurations, we use a Jacobian-based objective~\cite{ikdiffuser, yoshikawa}, replacing the raw Yoshikawa volume with the negative log-product of the constrained Jacobian singular values. This gives stronger gradients near singularities and remains well-defined when fewer than six task-relevant degrees of freedom are free.

\textit{Static stability:} For mobile manipulators, let $m(Q)$ denote the
center-of-mass margin inside the support polygon. Configurations whose margin
falls below a desired clearance $\delta$ can be penalized with \begin{equation} 
  E_\text{stability} (Q) = \beta \,\text{softplus} \left(\frac{\delta - m(Q)}{\beta}\right)
  \label{eq:stability}
\end{equation}
where $\beta$ controls the smoothness of the hinge. 

\textit{Behavior requests:} Let $\phi(Q)$ denote a differentiable solution
descriptor, such as base radius, bearing, or torso use. For an interval request
$\phi(Q) \in [a, b]$,  
\begin{equation*}
  E_\text{request} (Q) = \max(a - \phi(Q), 0)^2 + \max(\phi(Q)- b, 0)^2,
\end{equation*}
with analogous one-sided and target objectives. 

\subsubsection{Constraint-preserving guidance} Direct energy gradients can improve the requested objective while degrading IK accuracy. 
To steer primarily
along the set of solutions that already satisfy the task constraints, we compute a damped approximation to the guidance gradient's component in the local constraint tangent space~\cite{liegeois,nakamura_hanafusa}. 
Let $\chi(Q)$ be a vector of
the predicted end-effector position and orientation residuals, and let $G =
\partial \chi / \partial \hat{Y}$ be their Jacobian with respect to the output
features. Given an energy gradient $g = \nabla_{\hat{Y}} E$, its tangent
component is
\begin{equation*}
  g_{\parallel} = g - G^\top (GG^\top + \lambda I)^{-1} Gg.
\end{equation*}
The first term is the requested objective direction; the subtracted component
suppresses motion that would locally change the task constraints.  

Because the tangent approximation is only local, we also compute a damped
Gauss-Newton correction $r = - G^\top (GG^\top + \lambda I)^{-1}\chi$ and update
the output by \begin{equation*}
    \Delta \hat{Y} = -\eta g_{\parallel} + \kappa r,
\end{equation*} where $\eta$ is the guidance step size and $\kappa$ is the Gauss-Newton correction gain.

\subsubsection{Safeguards} We apply guidance only to free coordinates, and each sample's update is norm-clipped to $\|\Delta \hat{Y}\|_2 \leq
\Delta_{\max}$. We then pass the resulting sample through the same bounding
function $\mathcal{B}$ used for unguided sampling to preserve the joint-limit guarantee.  

\subsubsection{Redraw} A small fraction of generated samples may fall in the
tail of the FK-error distribution. Because FK is inexpensive to evaluate, we use
it to decide when an additional draw is worthwhile. For a decoded solution $Q$,
define the worst required-target position error
\begin{equation*} 
  e_\text{FK} (Q, c) = \max_{(\ell, e)} \|p_e (q_\ell) -p^\star_{\ell, e}\|_2.
\end{equation*}
If $e_\text{FK} > \tau$, the model draws once more from an independent latent
and keeps the candidate with lower FK error. We calibrate the threshold $\tau$
on held-out problems.

\subsubsection{Refinement} \label{sec:refinement} 
Learned samples provide informed, inexpensive
initializations for numerical optimization. We use a lightweight
Levenberg-Marquardt (LM)~\cite{levenberg, marquardt} optimizer, called a {\em refiner}, to reduce
residual task-space error. 
Because the sampler generates high-quality initializations quickly, we find that refining multiple samples with this lightweight optimizer is more cost-effective than invoking a more general-purpose IK solver.

\subsection{Conditional architecture}
\label{sec:method-model}

\begin{figure}[t]
  \centering
  \includegraphics[scale=0.24]{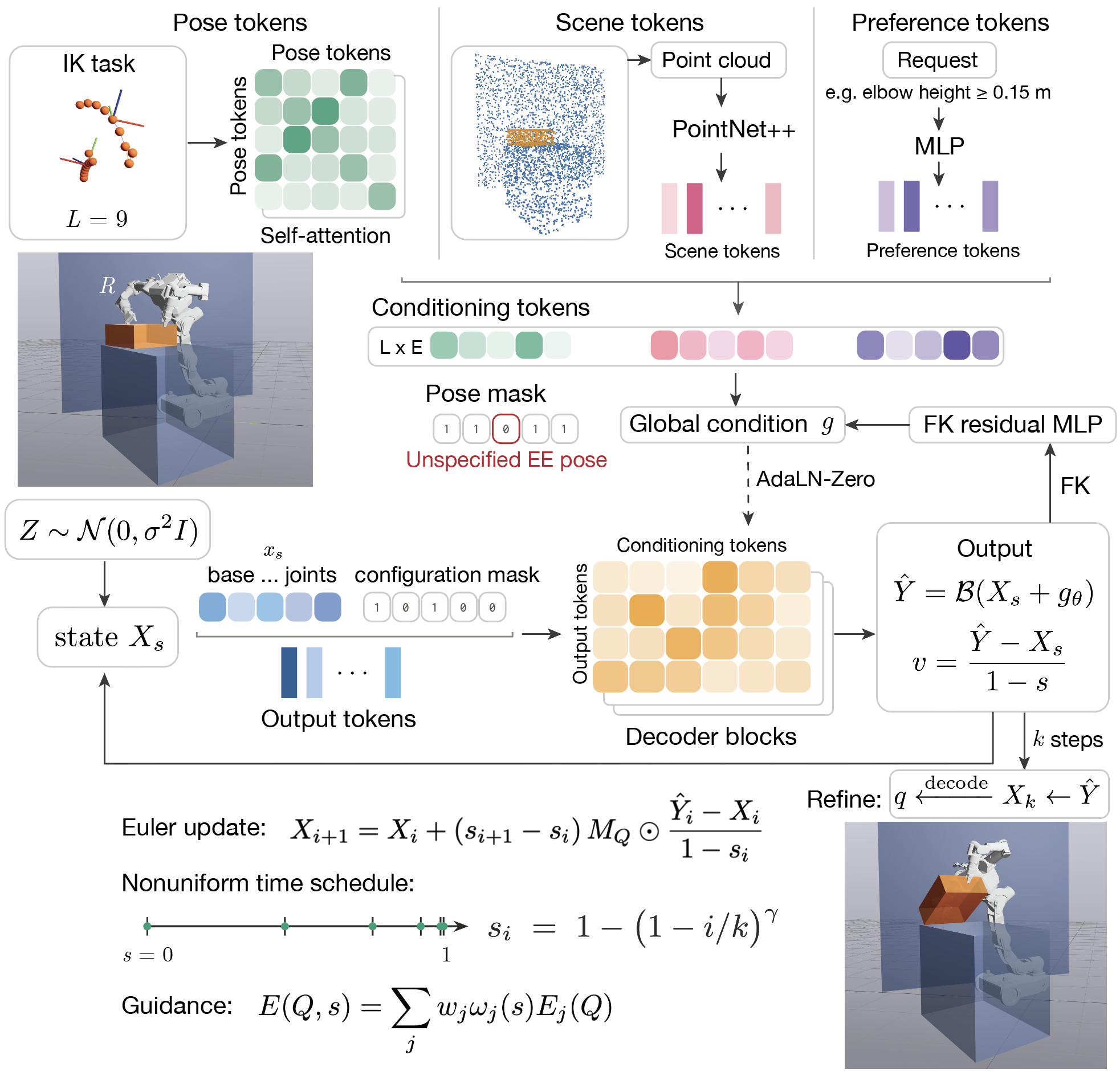}

  \caption{Conditional architecture. Task poses and scene geometry form conditioning tokens, while the current candidate is represented by base and joint tokens. The decoder combines these with FK feedback and optional solution requests.}
\label{fig:architecture}
\end{figure}

We implement $g_\theta$ with a conditional encoder-decoder Transformer, as illustrated in Fig.~\ref{fig:architecture}. Conditioning information enters the decoder through two paths: cross-attention over conditioning tokens and modulation by a global conditioning vector. 


\subsubsection{Task encoder} We use one token for each
(target, end-effector) pair, combining the target-pose features with target,
end-effector, and mask embeddings, and replace masked-out targets with a learned
null embedding. We process these tokens with self-attention and use attention pooling to produce a global conditioning vector.

\subsubsection{Solution decoder} The decoder receives one base token and one joint token per prediction target, each containing the current solution features and
known-coordinate mask. We embed the generation-time $s$ using a sinusoidal MLP
and add it to the global conditioning vector.
Decoder blocks apply self-attention across solution
tokens and cross-attention to the conditioning tokens, followed by SwiGLU~\cite{swiglu}
feed-forward layers, while the global conditioning vector modulates each block through AdaLN-Zero~\cite{dit}. Linear output heads map
the base and joint tokens back to their corresponding solution features. The
base has its own output head, while joint features may share one head or split
into separate output heads for predefined joint groups, such as the left arm,
right arm, and torso. 

\subsubsection{Scene conditioning} We represent scene geometry as a cloud of
4096 labeled points, marking each point as part of the environment or the manipulated object, in the same
canonical task frame as the robot and targets. A
PointNet++-style~\cite{pointnet++, pointnext} encoder reduces the 4096 input points to 256
local features and then 32 scene tokens, each representing local geometry around
a sampled centroid. Each token retains its centroid coordinates in the
canonical task frame to preserve the relative position of the task to the scene.
We append scene tokens to the conditioning tokens 
and add a pooled scene embedding to the global conditioning vector.

\subsubsection{Preference conditioning} We encode non-differentiable preferences such as standing posture or elbow branch, which are not amenable to guided sampling, as request tokens that specify a solution descriptor and the desired constraint on that descriptor, such as an ``elbow up" category or a minimum, maximum, or interval value. We append them to the conditioning tokens and add a pooled request embedding to the global conditioning vector. 

\subsubsection{Forward kinematics feedback} At each model evaluation, we compute 3D position and rotation errors for each end-effector relative to its target pose and include them as additional conditioning~\cite{graphdiffik}. We pass the residuals through an MLP and add them to the global conditioning vector.

\section{EXPERIMENTS}

 We evaluate whether \sysname{}'s learned conditional distribution can produce diverse candidates quickly across mobile-base-, multi-end-effector-, pathwise-, and scene-conditioned IK.  We also evaluate solution diversity and adherence to specified preferences. Unless otherwise specified, a sample is \emph{valid} if it satisfies the required task-space accuracy (0.1\,mm, 0.01\,rad), joint limits, and self- and world-collision constraints under a shared Drake mesh-collision evaluator. A query is \emph{successful@}$K$ if at least one of its $K$ samples is valid.  All results use held-out test problems.
 
 We measure time to a first valid solution (TTFV) by repeatedly drawing rounds of $K$ fresh samples, stopping at the first round containing a valid sample or after six rounds. Let $V$ denote the first successful round. For batch size $B$, let $t_{K, B}$ denote the wall-clock cost of one round (including refinement for our models). We report single-query $\text{TTFV}@6 = t_{K, 1} \mathbb{E}[\min(V, 6)]$ and additionally report amortized time per query as $(t_{K, B}/B) \cdot  \mathbb{E}[\min(V,6)]$. Because unsuccessful queries are capped at six rounds, low TTFV can reflect either fast success or fast failure and should therefore be interpreted alongside success rates; including failures avoids the bias of timing only the subset of queries a method solves. 
 All timings are measured on one NVIDIA RTX 4090 with an Intel i9-14900K.

For scene-conditioned comparisons, we tune cuRobo's batching and collision parameters for each setting, and score returned seeds individually rather than reporting only the best solution from a multi-seed solve to directly compare per-sample S@1 and multi-sample S@$K$. TTFV measurements are warmed and exclude Drake certification and cuRobo scene-loading/setup time.

\subsection{Training}
For scene-free point IK, we use $10^{7}$ configurations per robot. For scene-conditioned and pathwise tasks, we use approximately 600{,}000 grasp pose examples across 27 MolmoSpaces kitchens~\cite{molmo} and 500{,}000 examples per task family across $\approx$2{,}500 procedurally generated environments.

All models use 2 encoder layers and 6 decoder layers with $d_\text{model} = 256$ ($\approx$12M parameters) or 512 ($\approx$45M);
free-space models use the larger backbone, while more
constrained task families use the smaller one. We train with AdamW~\cite{adamw} (lr $10^{-4}$, weight
decay $10^{-4}$, warmup-cosine schedule), bf16, and EMA over weights (decay
0.999).  

\subsection{Scene-free generalized IK} 

We first evaluate whether a single modeling and training recipe can produce accurate IK solutions across fixed-base, floating-base, whole-body, and multi-end-effector settings. Table~\ref{tab:ik-main} shows that \sysname{} achieves comparable or better accuracy than other learned IK methods while using few network evaluations (NFEs). We also evaluate distributional fidelity: following IKFlow's MMD evaluation against TRAC-IK~\cite{trac-ik}, our model obtains 0.0302 on FR3, comparable to IKFlow's reported 0.0306 on Panda. On scene-free point IK, cuRobo achieves lower latency under large GPU batches, while our sampler attains higher per-seed convergence.

\begin{table}[t]
\centering
\begin{threeparttable}
\caption{Scene-free point IK. Robots are specified as (number of end-effectors, degrees of freedom [DoF]). Position and rotation errors are mean $\pm$ standard error over 100 samples for each of 10,000 targets. NFE includes redraws. Reported baselines are from the respective papers (standard errors are estimated assuming 100 samples). Conv. measures per-sample convergence when seeding cuRobo with model samples, requiring position error below 0.1\,mm and rotation error below 0.01\,rad.}

\label{tab:ik-main}
\scriptsize
\setlength{\tabcolsep}{3pt}
\renewcommand{\arraystretch}{0.9}
\begin{tabular}{@{}llccccc@{}}
\toprule
\textbf{Robot} &
\textbf{Method} &
\makecell{\textbf{Self-coll.}\\ (\%) $\downarrow$} &
\makecell{\textbf{Pos. err.}\\ (mm) $\downarrow$} &
\makecell{\textbf{Rot. err.}\\ (deg) $\downarrow$} &
\makecell{\textbf{NFE}\\ $\downarrow$} &
\makecell{\textbf{Conv.}\\ (\%) $\uparrow$} \\
\midrule

\multirow{2}{*}{\makecell[l]{FR3 (1, 7)}}
& IKD~\cite{ikdiffuser}
& 0.49
& 4.90 $\pm$ 0.284
& 1.05 $\pm$ 0.062
& 5
& 98.47 \\
& Ours
& 0.52
& 4.73 $\pm$ 0.275
& 0.78 $\pm$ 0.057
& 4.1
& 98.50\\
\midrule

\multirow{2}{*}{\makecell[l]{LEAP (4, 22) \\6DoF-base}}
& IKD~\cite{ikdiffuser}
& 2.85
& 1.73 $\pm$ 0.072
& 0.93 $\pm$ 0.033
& 5
& 80.95 \\
& Ours
& 1.18
& 1.14 $\pm$ 0.049
& 0.80 $\pm$ 0.034
& 2.4
& 95.81 \\
\midrule

\multirow{2}{*}{\makecell[l]{G1 (4, 29)}}
& IKD~\cite{ikdiffuser}
& 6.35
& 5.31 $\pm$ 0.156
& 1.48 $\pm$ 0.049
& 5
& 96.96\\
& Ours
& 1.97
& 3.44 $\pm$ 0.159
& 0.80 $\pm$ 0.042
& 4.6
& 98.30\\
\midrule

\multirow{2}{*}{\makecell[l]{R1Pro (2, 18) \\ fixed-base}}
& GD~\cite{graphdiffik}
& --
& 5.29 $\pm$ 0.386
& 0.59 $\pm$ 0.035
& 100
& --\\
& Ours
& 0.05
& 2.73 $\pm$ 0.242
& 0.32 $\pm$ 0.033
& 9.2
& 99.97\\

\bottomrule
\end{tabular}
\end{threeparttable}
\end{table}

\subsection{Constraints required by planning}

Planning imposes structure beyond point IK: solutions must account for surrounding geometry, remain coherent across trajectories, and reflect task-specific preferences. We test each of these capabilities.

\textbf{Scene geometry.} We evaluate on held-out scenes spanning both in-distribution kitchens and out-of-distribution room types (bathrooms, bedrooms, and living rooms). We test whether scene-conditioned generation uses the observed geometry 
by evaluating the same model with the correct scene cloud, no scene input, and a cloud shuffled from 
another scene. We additionally evaluate partial observability from a single pre-move depth image, 
comparing models trained on full and partial scene clouds against cuRobo using voxelized 
observations. For cuRobo, we use the best-performing batching, collision parameters, and scene 
representation found for each observation setting. Table~\ref{tab:scene-coll} summarizes these 
results, and Fig.~\ref{fig:scenes} shows example scenes.
The same interface also lets a planner introduce temporary exclusion regions as additional scene geometry.

\begin{table}[t]
\centering
\caption{Scene-aware point IK on RB-Y1. Results use 1024 grasp problems from 252 held-out iTHOR and ProcTHOR scenes. Models are trained on full point clouds (full), partial point clouds (partial), or no scene data (blind). 
Ours uses a 32-NFE model with our refiner unless otherwise specified; cuRobo uses cuboids for full scenes and 2.5 cm voxels with unobserved space free for partial scenes. All samples are evaluated against the complete ground-truth scene. Columns specify success@1, success@8, and world-collision-free@1. TTFV uses rounds of $K{=}8$ samples.}
\label{tab:scene-coll}
\scriptsize
\setlength{\tabcolsep}{3pt}
\renewcommand{\arraystretch}{0.9}

\begin{tabular}{@{}llcccccc@{}}
\toprule
\textbf{Method} & \textbf{Train} & 
\textbf{Test} &
\makecell{\textbf{S@1} \\ (\%)} &
\makecell{\textbf{S@8}\\ (\%)} &
\makecell{\textbf{WF@1}\\ (\%)} &
\multicolumn{2}{c}{\makecell{\textbf{TTFV} (ms) \\ B1\quad B32}} \\
\midrule

Ours     & full        & full            & 80.9 & 98.2 & 84.7 & 66.4 & 5.6 \\
Ours (1-NFE)   & full   & full            & 71.2 & 96.8 & 75.6 & 12.1 & 2.4 \\
Ours   & full         & none                   & 16.7 & 52.3 & 17.6 & 156 & 11.8 \\
Ours  & full             & wrong cloud            & 20.7 & 40.7 & 22.7 & 225 & 18.9 \\
Ours (blind)      & none      & none                   & 14.6 & 48.6 & 16.5 & 171 & 18.4 \\
Ours (raw) + cuRobo    & full & full (cuboids)       & 90.6 & 93.3 & 95.4 & 173 & 92.2 \\
cuRobo        & none     & full (cuboids)               & 29.0 & 58.1 & 59.9 & 217 & 188 \\
cuRobo             & none    & none               & 7.8 & 36.9 & 9.9 & 44.2 & 5.3 \\

\midrule

Ours  & full          & partial          & 53.3 & 82.2 & 55.5 & 102 & 8.3 \\
Ours  & partial   & partial          & 61.4 & 90.1 & 64.4 & 83.1 & 6.9 \\
Ours (raw) + cuRobo & partial & partial (voxels)  & 57.7 & 81.6 & 68.8 & 195 & 74.1 \\
cuRobo       & none           & partial (voxels)       & 12.0 & 52.1 & 25.5 & 132 & 98.8 \\

\bottomrule
\end{tabular}
\end{table}

\textbf{Path IK.} We evaluate whether \sysname{} can generate coherent multi-target realizations that remain feasible across an entire task while keeping the mobile base fixed. We consider door and drawer manipulation, pushing, bimanual lifting, flip-up, and under-table pickup tasks on Rainbow RB-Y1. Models are trained on procedurally generated instances of the same task families, with randomized task motions, object geometry, and surrounding obstacles.

For path tasks, the defined validity criterion is applied at every specified target. We additionally densify the configuration-space interpolation between targets and require all intermediate states to remain collision-free; bimanual and under-table tasks also require quasi-static stability. The cuRobo baseline propagates all valid first-target solutions from 32 seeds as independent fixed-base, warm-started IK continuations. Our solver receives the applicable scene geometry and task constraints (a linear-hinge version of the stability cost in Eq.~\ref{eq:stability}). Table~\ref{tab:path-ik} reports success@$K$ and TTFV under these criteria.

\begin{table}[t]
\centering
\begin{threeparttable}
\caption{Path IK on RB-Y1. Tasks are specified as (number of end-effectors, path length(s)). Ours uses 1-NFE reflow students and our refiner. Each task family uses 150 problems from 150 distinct scenes with 32 draws per problem. TTFV uses rounds of $K{=}32$ samples.} 

\label{tab:path-ik}
\scriptsize
\setlength{\tabcolsep}{6pt}
\renewcommand{\arraystretch}{0.9}
\begin{tabular}{@{}llcc@{}}
\toprule
\textbf{Task} &
\textbf{Method} &
\makecell{\textbf{Success@32} (\%)} &
\makecell{\textbf{TTFV},\,B1 (ms)} \\
\midrule

\multirow{2}{*}{\makecell[l]{Door/drawer (1, $L{=}$3, 5)}}
& cuRobo 
& 89.3
& 170.5 \\
& Ours
& 100.0
& 22.7 \\
\midrule

\multirow{2}{*}{\makecell[l]{Pushing (1, $L{=}$2, 5)}}
& cuRobo 
& 92.0
& 167.3 \\
& Ours
& 98.7
& 20.2 \\
\midrule

\multirow{2}{*}{\makecell[l]{Bimanual lift (2, $L{=}$5)}}
& cuRobo 
& 42.7
& 940.0 \\
& Ours
& 92.7
& 36.1 \\
\midrule

\multirow{2}{*}{\makecell[l]{Flip-up (2, $L{=}$9)}}
& cuRobo 
& 22.0
& 1067.2 \\
& Ours
& 98.0
& 31.2 \\
\midrule

\multirow{2}{*}{\makecell[l]{Under-table (2, $L{=}5$)}}
& cuRobo 
& 60.7
& 575.9 \\
& Ours
& 93.3
& 23.2 \\
\bottomrule
\end{tabular}
\end{threeparttable}
\end{table}

We next test whether conditioning on the full path improves the choice of the first configuration. We compare cuRobo samples given only $H_1$ with our samples conditioned on $H_{1:L}$, then apply the same fixed-base cuRobo continuation from each valid $q_1$. We call a valid $q_1$ \emph{extendable} if its continuation passes the full-path validity checks of Table~\ref{tab:path-ik}. As shown in Table~\ref{tab:path-extend}, full-path-conditioned \sysname{} yields 58.2\% extendable starts on average, versus 24.3\% for cuRobo given only $H_1$, with improvements on every task.

\begin{table}
\centering
\caption{Fraction (\%) of valid first-target configurations that admit a valid cuRobo continuation, evaluated on 100 tasks per task family.}
\label{tab:path-extend}
\footnotesize
\setlength{\tabcolsep}{4pt}
\renewcommand{\arraystretch}{1.0}
\begin{tabular}{@{}lcccccc@{}}
\toprule
\textbf{$q_1$ source} & \textbf{Push} & \textbf{Door} & \textbf{Drawer} & \textbf{Lift} & \textbf{Flip} & \textbf{Under-table}\\
\midrule
cuRobo, $H_1$ only & 26.0 & 18.5 & 47.4 & 41.2 & 5.4 & 7.3 \\
Ours, full path & 76.2 & 56.2 & 87.0 & 52.9 & 29.1 & 47.9 \\
\bottomrule
\end{tabular}
\end{table}

Finally, we test whether the model learns structure that transfers across manipulation families rather than requiring a separate specialist for each. With all families present, the unified model performs comparably to task-specific specialists, with success@32 rates within 2.2 percentage points. We then hold out individual families entirely from training. Performance remains high on pushing (97.3\%, versus 98.7\% for the specialist in Table~\ref{tab:path-ik}) and transfers substantially to flip-up (76.7\%, versus 98.0\%), despite flip-up requiring trajectory lengths not seen during training. In contrast, performance on held-out under-table pickup falls to 0\%. These results suggest that the model transfers task-space and arm-motion structure across families, but does not extrapolate to qualitatively new scene interactions (reaching into a confined volume) absent from training.

\subsection{Preferences}
\begin{figure}[t]
  \centering
  \includegraphics[scale=0.173]{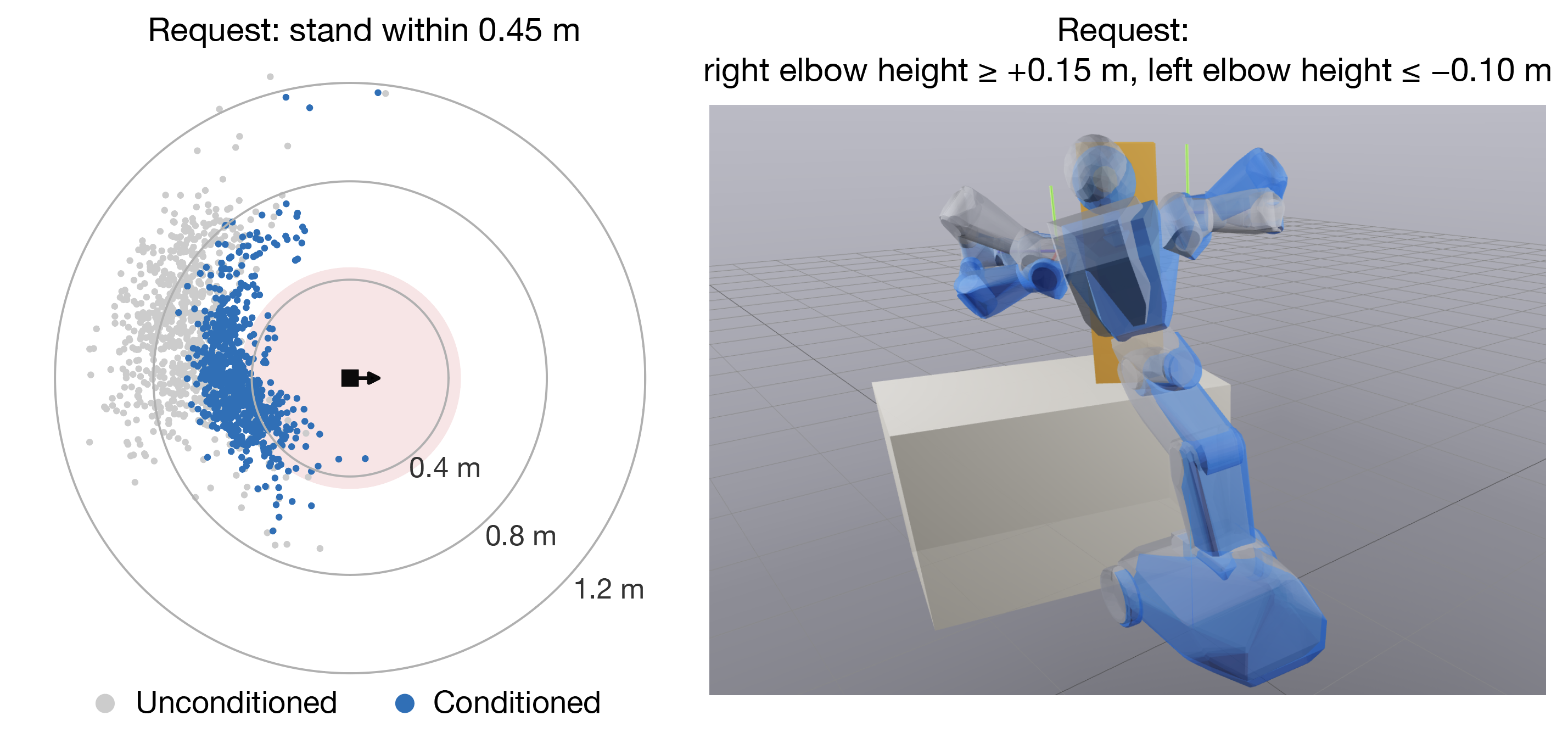}

  \caption{Preference-conditioned IK in the RB-Y1 bimanual lift task. Left: requesting a base stand-off ${\leq}0.45$\,m shifts generated base placements inward relative to unconditional samples, where stand-off is measured from the anchor end-effector target (black box) to the base axle. 
  Right: requesting right-elbow height ${\geq }{+}0.15$\,m and left-elbow height ${\leq} {-}0.10$\,m yields an asymmetric posture, where elbow height is the world-frame vertical offset from the shoulder.}
\label{fig:preference-conditioning}
\end{figure}

We test whether we can directly condition the generated solution distribution on preferences without requiring repeated rejection or retraining. Fig.~\ref{fig:preference-conditioning} shows that preference conditioning shifts the sampled solution distribution toward requested regions relative to unconditional sampling. 

For differentiable objectives not specified during training, we apply inference-time guidance. Table~\ref{tab:guidance} reports results on the G1 robot for manipulability, warm-start proximity, and static stability, measured respectively by relative increase in mean Yoshikawa manipulability~\cite{yoshikawa}, range-normalized joint distance to the supplied prior, and relative reduction in static-stability violation rate. For the stability evaluation, we constrain both feet to the specified end-effector poses, so the support polygon remains fixed. 
We construct the polygon as the convex hull of the foot contacts, eroded inward by 3 cm, and count a violation when the projected whole-body center of mass lies outside it. Guided and unguided evaluations use the same latent draws. Guidance improves all requested objectives while preserving task-space accuracy: across objectives, mean position error decreases by 0.28--2.39\,mm and mean rotation error decreases by 0.14--0.47$^\circ$, consistent with the Gauss-Newton correction during guidance.

\begin{table}[t]
\centering
\begin{threeparttable}
\caption{Inference-time guidance on G1; 10{,}000 targets ${\times}$ 100 samples.}
\label{tab:guidance}
\scriptsize
\setlength{\tabcolsep}{5pt}
\renewcommand{\arraystretch}{0.92}
\begin{tabular}{@{}lccc@{}}
\toprule
&
\makecell{\textbf{Manipulability gain} (\%) $\uparrow$} &
\makecell{\textbf{Distance to} \\ \textbf{prior} (\%) $\downarrow$} &
\makecell{\textbf{Stability violation} \\ \textbf{reduction} (\%) $\uparrow$}\\
\midrule

IKDiffuser~\cite{ikdiffuser} & 24.69 & 4.95 & -- \\
Ours       & 28.42 & 1.59 & 68.74 \\
\bottomrule
\end{tabular}
\end{threeparttable}
\end{table}

\section{CONCLUSION}
We present \sysname, a generative modeling approach to generalized IK. Experiments show that our model significantly increases solve rates of optimization-based methods in manipulation tasks within novel scenes while allowing flexible control over the sampling distribution via train-time conditioning and inference-time guidance. Our system batches queries across multiple task families and distinct scenes; a one-step sampler followed by lightweight refinement reduces solve time by up to 34$\times$ relative to batched numerical optimization. Notably, a unified model matches the performance of task-specific specialists, and we observe instances of generalization to unseen task families, suggesting that future work may extend \sysname{} to new manipulation tasks without requiring expert-solver supervision for every task family.

\section*{Acknowledgment}

We gratefully acknowledge support from AFOSR grant FA9550-22-1-0249; from ARO grant W911NF-26-1-A329; from ONR MURI grants N00014-22-1-2740 and N00014-24-1-2603; from the MIT Siegel Family Quest for Intelligence; and from the Robotics and AI Institute.

\noindent
\textbf{AI disclosure:} Claude Code assisted with code completion and debugging; all code was author-reviewed.


\bibliographystyle{IEEEtran}
\bibliography{references}

\appendix

\subsection{Data and optimization}
\label{app:train}
We split scene-conditioned datasets into 80\% training and 20\% validation sets by scene. Training solutions pass the validity checks in Appendix~\ref{app:eval}, with task-space tolerances of 2\,mm and 0.02\,rad at each target. All path-training solutions additionally track the interpolated target path within 5\,mm and 0.04\,rad.

We use batches of 1024 solutions, reduced to 512 for bimanual and under-table tasks. Table~\ref{tab:hparams} lists settings that supplement the architecture and optimization details in the main text. All models train within 10 hours on a single NVIDIA GH200 GPU (96\,GB HBM3). 

\begin{table}[h]
\centering
\caption{Selected training hyperparameters.
Teacher auxiliary-loss weights are relative to the
flow-matching loss.}
\label{tab:hparams}
\footnotesize
\setlength{\tabcolsep}{6pt}
\renewcommand{\arraystretch}{1.08}
\begin{tabular}{@{}lll@{}}
\toprule
\textbf{Parameter} & \textbf{Models} & \textbf{Value} \\
\midrule
\multicolumn{3}{@{}l}{\textit{Shared}} \\
Soft-bound margin $\mu$ & All & $2^\circ$ \\
$S^1$ weight & All & 0.05 \\
Minimum time $\epsilon$ & Except under-table & 0.01 \\
                        & Under-table & 0.02 \\
\midrule
\multicolumn{3}{@{}l}{\textit{Teacher training}} \\
Peak learning rate & Except under-table & $10^{-4}$ \\
                   & Under-table & $5 \times 10^{-5}$ \\
\midrule
\multicolumn{3}{@{}l}{\textit{Teacher fine-tuning}} \\
Peak learning rate & All fine-tuned models & $2 \times 10^{-5}$ \\
Scene-collision weight & Flip-up, under-table & 0.25 \\
                       & Full-cloud point IK & 0.10 \\
                       & Partial-cloud point IK & 2.8 \\
Self-collision weight & Flip-up, under-table & 0.25 \\
                      & Partial-cloud point IK & 0.70 \\
Stability weight & Flip-up, under-table & 0.25 \\
Task-space counterweight & Model-dependent & 0.005--0.02 \\
\midrule
\multicolumn{3}{@{}l}{\textit{Student distillation}} \\
Scene-collision weight & Flip-up, under-table & 0.02 \\
Stability weight & Flip-up, under-table & 0.2 \\
\bottomrule
\end{tabular}
\end{table}

\subsection{Sampler-aware fine-tuning} 
We differentiate through the inference-time Euler sampler and penalize the decoded output's task-space error: 
\begin{equation*}
  \mathcal{L}_{\mathrm{roll}}
  = \frac{e_{\mathrm{pos}}^2}{\sigma_p^2}
  + \frac{e_{\mathrm{rot}}^2}{\sigma_r^2},
\end{equation*}
where $\sigma_p = 5$\,mm and $\sigma_r = 1^\circ$. 

Before distillation, we fine-tune the flip-up and under-table teachers for 15k steps with scene-collision, self-collision, and stability penalties; the flip-up stability penalty adds the carried object's mass at its commanded pose. We evaluate these penalties
at $s{=}0.97$ along the 32-step sampling schedule and retain
a task-space counterweight to preserve target accuracy.

Scene-collision penalties use a 5\,mm margin, 1\,cm scale,
and 2\,cm cap. The stability penalty uses
a 2\,cm support-polygon inset, 1\,cm scale, and 5\,cm cap.

\subsection{Reflow distillation}
For each training solution, we draw four latents and integrate the teacher for 32 steps to construct latent--endpoint pairs. We initialize students with teacher EMA weights and train only at $s{=}0$, matching one-step deployment, by regressing the predicted endpoint onto the teacher endpoint. 

\subsection{Evaluation details}
\label{app:eval}
Validity follows Sec. V; here we specify the exact thresholds. Joint limits hold to within $10^{-6}$, and robot links keep at least 1 mm clearance from each other and from the scene. For path tasks, the base is identical across targets, and the joint-space interpolation is checked at 17 evenly spaced configurations, which include every target; joint limits and collision apply at each. 

\end{document}